\documentclass[letterpaper]{article} 
\usepackage{aaai2027}  
\usepackage[hyphens]{url}  
\usepackage{graphicx} 
\usepackage{natbib}  
\usepackage{caption} 
\usepackage{algorithm}
\usepackage{algorithmic}

\usepackage{newfloat}
\usepackage{listings}
\DeclareCaptionStyle{ruled}{labelfont=normalfont,labelsep=colon,strut=off} 
\floatstyle{ruled}
\newfloat{listing}{tb}{lst}{}
\floatname{listing}{Listing}

\usepackage{booktabs}

\usepackage{algorithm}
\usepackage{algorithmic}

\usepackage{comment}
\usepackage[hyphens]{url}
\usepackage{pgfplots}
\pgfplotsset{compat=1.18}
\usepackage{pgfplotstable}
\usepgfplotslibrary{groupplots,statistics}

\usepackage{multirow}
\usepackage{tikz}
\usepackage{xspace}

\usepackage{pifont}

\usepackage[table]{xcolor}
\usepackage{amsmath,amssymb}
\usepackage{siunitx}

\newcommand{\ours}{\texttt{EB-CaP}\xspace}

\definecolor{pinkNonTTA}{RGB}{255,248,251}
\definecolor{pinkTTA}{RGB}{255,240,246}
\definecolor{blueTTA}{RGB}{240,246,255}

\definecolor{coral}{RGB}{255,127,80}
\definecolor{teal}{RGB}{0,128,128}
\definecolor{violet}{RGB}{138,43,226}
\definecolor{gold}{RGB}{255,191,0}

\definecolor{biovidcolor}{RGB}{213,94,0}
\definecolor{stresscolor}{RGB}{0,114,178}
\definecolor{bahcolor}{RGB}{0,158,115}

\newcommand{\biovid}{\texttt{BioVid}\xspace}
\newcommand{\stressid}{\texttt{StressID}\xspace}
\newcommand{\bah}{\texttt{BAH}\xspace}

\newcommand{\war}{\texttt{WAR}\xspace}
\newcommand{\fonescore}{\texttt{F1}\xspace}

\title{Test-Time Adaptation with Online Personalized Energy-Based Cache for \\ Fine-Grained Video Expression Recognition}
\author{
    Masoumeh Sharafi\textsuperscript{\rm 1},
    Muhammad Osama Zeeshan\textsuperscript{\rm 1},
    Soufiane Belharbi\textsuperscript{\rm 1},
    Alessandro Lameiras Koerich\textsuperscript{\rm 2},
    Marco Pedersoli\textsuperscript{\rm 1},
    Eric Granger\textsuperscript{\rm 1}
}
\affiliations{
    \textsuperscript{\rm 1}LIVIA, Department of Systems Engineering, 
    École de technologie supérieure, Montreal, Canada\\
    \textsuperscript{\rm 2}LIVIA, Department of Software and IT Engineering, 
    École de technologie supérieure, Montreal, Canada\\
    \{masoumeh.sharafi.1,muhammad-osama.zeeshan.1\}@ens.etsmtl.ca\\
    \{soufiane.belharbi,marco.pedersoli,alessandro.koerich,eric.granger\}@etsmtl.ca
}

\begin{document}

\maketitle

\begin{abstract}
Facial expression recognition (FER) in videos remains challenging because models must identify subtle temporally evolving affective states that vary significantly across target individuals. Although vision-language models provide transferable visual-semantic representations, models trained on subject-independent source data often degrade under subject-specific distribution shifts at inference time. State-of-the-art test-time adaptation (TTA) methods typically optimize models during inference, increasing computational cost and latency. Cache-based approaches avoid parameter updates but typically require accumulating sufficient target samples to construct reliable class prototypes. This is difficult at the beginning of adaptation and when some classes are rarely observed. To alleviate these limitations, existing methods may store source prototypes, but these are not personalized to the current target subject.
This paper introduces Energy-Based Cache Personalization (\ours), a subject-based online TTA method for video FER that samples class-specific prototypes personalized to each target video on-the-fly. Unlike existing cache-based methods, \ours does not require observing and accumulating large amounts of target data or storing diverse source prototypes across subjects. Instead, it relies on a lightweight energy-based model (EBM) to sample class-wise prototypes from the current unlabeled video and populate a personalized cache online. The energy function relies only on the pretrained CLIP model, where the similarity between the visual embedding of the target video and the class text embeddings guides the energy-based sampling process. In parallel, positive and negative caches store reliable and uncertain target embeddings, respectively. An adaptive entropy gating follows the evolving confidence distribution to control cache updates, while a diversity gate prevents redundant samples from dominating the memory. Predictions are refined by combining cache-derived scores with the current CLIP scores. 
Experiments on 3 challenging video datasets for video FER -- \biovid, \stressid, and \bah~-- indicate that \ours can outperform state-of-the-art TTA methods, while maintaining low computational and memory overhead. Our code is publicly available at \url{https://github.com/MasoumehSharafi/EB-CaP}.
\end{abstract}


\begin{figure}[t!]
  \centering
  \includegraphics[width=0.86\linewidth]{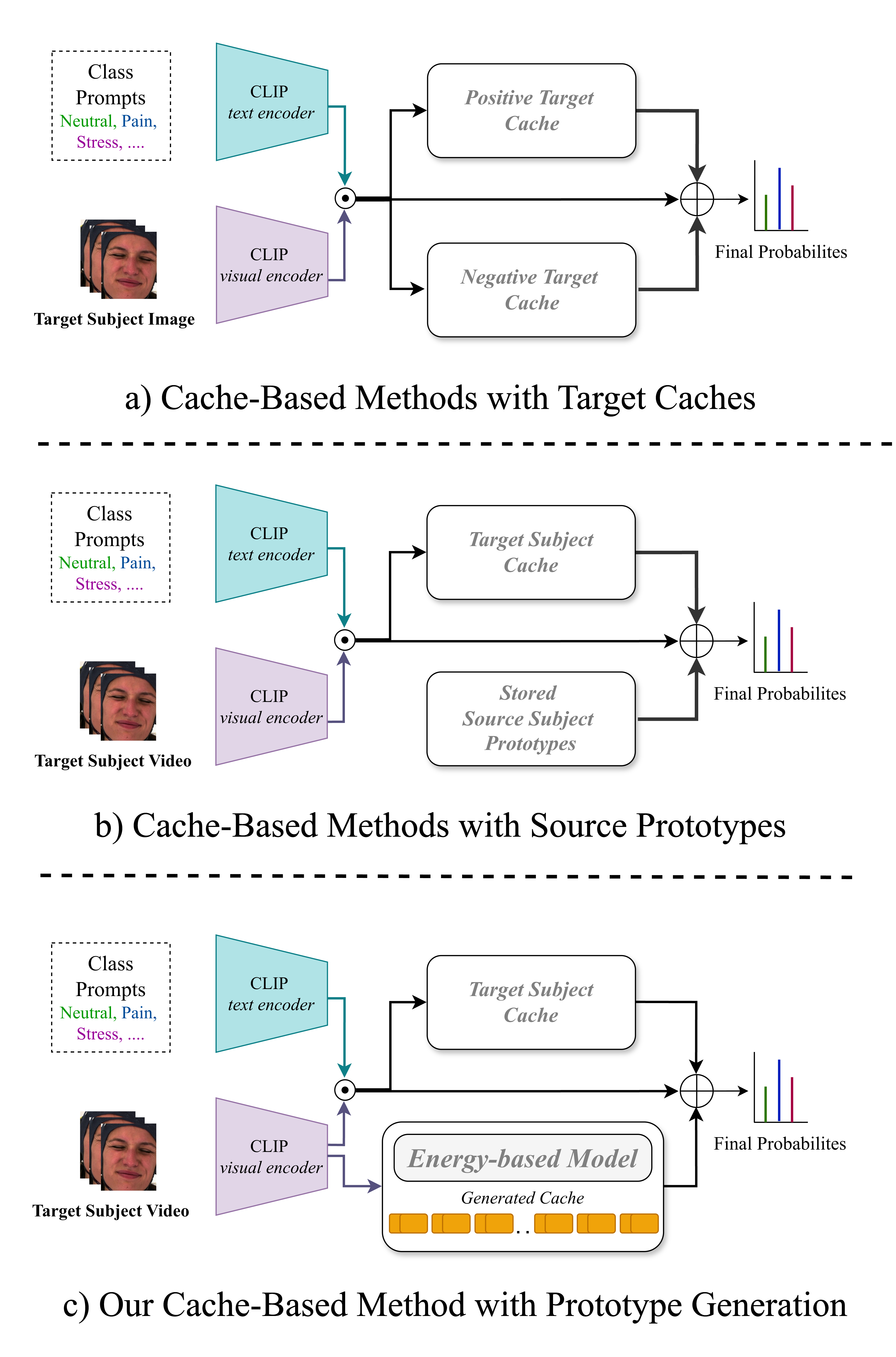}
  \caption{Cache-based TTA methods from target subject data. \textbf{(a)} State-of-the-art methods with dynamic positive and negative target caches may accumulate confident but incorrect pseudo-labels~\cite{karmanov2024efficient}. \textbf{(b)} A prototype-based method mitigates this issue by constructing prototypes from source domain samples with ground-truth labels and storing them at test time~\cite{sharafi2026test}. \textbf{(c)} In contrast, \ours samples subject-specific, class-wise prototypes online from each unlabeled target video using an EBM.}
  \label{fig:Motivation}
\end{figure}

\section{Introduction}

Fine-grained FER in videos is challenging as closely related affective states may differ only in subtle facial movements, expression intensity, or temporal evolution. These cues also vary across individuals, causing the same state to appear differently across subjects while different states may exhibit similar facial patterns. Consequently, models trained on fixed source subjects often generalize poorly to unseen individuals. This limitation is particularly important in many applications including human--computer interaction~\cite{pu2023convolutional}, health monitoring~\cite{gayamorey2025deeplearningbasedfacialexpression}, and the assessment of pain~\cite{sharafi2025disentangled, sharafi2025personalized, sharafi2026test,zeeshan2024subject, zeeshan2025musaco, zeeshan2025progressive}, depression~\cite{de2020deep}, and stress~\cite{calvo2010affect}. Reliable fine-grained video FER, therefore, requires representations that can capture subtle expression cues, temporal dynamics, and subject-specific variability.

VLMs, and in particular CLIP models~\cite{radford2021learning}, provide a promising foundation for fine-grained FER by transferring semantic knowledge learned from large-scale visual--text data~\cite{yu2022coca,li2021supervision,li2023blip}. Recent CLIP-based models for FER have further improved recognition through task-specific prompting, temporal modeling, and richer textual supervision~\cite{foteinopoulou2024emoclip,zhao2023prompting,zhao2025enhancing}. However, these approaches mainly learn subject-independent representations and may degrade when the facial behavior, appearance, or recording conditions of an unseen subject differ from those observed during source training~\cite{shu2023clipood,tu2023closer}. This limitation is especially important in fine-grained FER, where small subject-specific variations can alter predictions between closely related affective states. TTA can address this issue by personalizing a model to target subjects based on unlabeled videos captured at inference time. State-of-the-art CLIP-based TTA methods are broadly optimization-based or cache-based. Optimization-based methods update prompts, normalization statistics, or lightweight model components using objectives such as entropy minimization and prediction consistency~\cite{shu2022test,feng2023diverse,abdul2023align,yoon2024c,zhang2024robust}; however, repeated online optimization increases inference costs and may cause unstable adaptation when target data are ambiguous or incorrectly predicted.

Cache-based TTA offers a more efficient alternative, where target embeddings and their pseudo-labels are stored and retrieved to refine subsequent predictions~\cite{karmanov2024efficient,zhang2024dual,huang2025cosmic,nguyen2025adaptivecacheenhancementtesttime,liang2025advancing,zhu2025dynamic,zhai2025mitigating,chen2025multi,chen2025paf,guanstatistics}. Since model parameters remain fixed, these methods are suitable for real-time applications. However, they often require sufficient target samples to form reliable class prototypes, creating a cold-start problem when classes are rare or initially absent. Some methods store source prototypes~\cite{sharafi2026test}, but these require extra storage and are not personalized to the current subject. The diversity of observed samples limits target caches. In video FER, highly correlated temporal windows can cause redundant high-confidence embeddings to dominate the cache, while incorrect pseudo-labels may propagate across ambiguous transitions. An effective cache-based method should therefore construct reliable, diverse, and personalized prototypes without requiring large amounts of accumulated target data.

This paper introduces Energy-Based Cache Personalization (\ours), a cache-based TTA method for video FER that operates without model optimization at inference time. It samples online using an energy-based model (EBM)~\cite{duvenaud2020your} to generate class-wise prototypes from each unlabeled target video. As illustrated in Figure~\ref{fig:Motivation} contrasts three cache construction strategies: target-only methods accumulate positive and negative target embeddings but suffer from cache cold start (Fig.~\ref{fig:Motivation}(a)); prototype-based methods store prototypes derived from labeled source data, providing more reliable class references when target caches contain high-confidence but incorrectly pseudo-labeled samples (Fig.~\ref{fig:Motivation} (b)); and \ours generates subject-specific, class-wise prototypes online from each unlabeled target video without storing source data or labels (Fig.~\ref{fig:Motivation}(c)). The sampling process starts from the current video representation and is stochastic, enabling the generation of multiple diverse prototypes for each class. To implement this sampling strategy, \ours uses a lightweight class-conditional EBM whose energy is guided by the similarity between the target visual embedding and the CLIP class text embeddings. Generated prototypes are stored in a personalized sampled cache that is refreshed for every incoming video. In parallel, a positive cache stores reliable target embeddings, while a negative cache retains uncertain or class-conflicting embeddings as negative evidence. These target caches are maintained across videos of the same subject and reset for each new subject. Their updates are controlled by an adaptive entropy gate and a diversity gate that limits redundant samples. During inference, cosine similarities between the incoming video and all three caches are converted into class scores and combined with the base CLIP scores to obtain the final prediction.

The main contributions of this work are summarized as follows. \textbf{(1)} \ours: an online strategy that rapidly samples diverse class-wise embeddings personalized to each target video. It addresses the cache cold start problem without accumulating target data or retaining source prototypes. A lightweight EBM guided by CLIP visual-text similarity enables this sampling. \textbf{(2)} Adaptive entropy and diversity gates for reliable cache construction. The entropy gate updates the caches based on each incoming video, while the diversity gate selects informative and non-redundant embeddings, reducing noisy pseudo-labels and repeated cache entries. \textbf{(3)} An extensive set of experiments on three challenging fine-grained video FER benchmarks -- \biovid~\cite{walter2013biovid}, \stressid~\cite{chaptoukaev2023stressid}, and \bah~\cite{gonzalez2025bah} -- show that \ours outperforms state-of-the-art optimization-based and cache-based TTA methods while maintaining low computational and memory overhead.

\section{Related Work}

\noindent\textbf{Vision-Language Models for FER.}
CLIP enables prompt-based recognition by matching visual features with semantic text representations~\cite{radford2021learning}, making it suitable for FER. However, facial expressions are fine-grained and affected by subject identity, temporal dynamics, and recording conditions. Recent CLIP-based FER methods improve textual guidance and temporal modeling through expression descriptions, language priors, and parameter-efficient adaptation~\cite{li2024cliper,zhao2023prompting,foteinopoulou2024emoclip,zhao2025enhancing,ma2025mpafer}. Nevertheless, they mainly focus on training-time adaptation and prompt design, leaving test-time personalization for video FER less explored.

\noindent\textbf{Test-Time Adaptation.}
TTA adapts pretrained models to unlabeled test data under distribution shift. Early methods use self-supervised objectives or entropy minimization, while later approaches improve robustness under noisy, small-batch, and non-stationary streams through selective updates, teacher--student learning, and memory mechanisms~\cite{sun2020test,zeeshan2026clip,wang2022continual,yuan2023robust}. However, repeated updates may accumulate errors and cause model drift. For VLMs, TPT, and PromptAlign adapt prompts or align test-time feature statistics, while other methods use prompt ensembling, weight averaging, or CLIP-specific objectives~\cite{shu2022test,abdul2023align,osowiechi2024watt,lafon2025cliptta}. These approaches require forward and backward passes and remain sensitive to noisy predictions and optimization settings.

\noindent\textbf{Cache-Based TTA.}
Cache-based TTA keeps the backbone frozen and adapts predictions through retrieval from a memory~\cite{karmanov2024efficient}. CLIP-Adapter, Tip-Adapter, and CODER improve CLIP predictions using adapters, key-value caches, or neighborhood retrieval~\cite{gao2024clip,zhang2022tip,yi2024leveraging}. Recent methods strengthen cache reliability through selective updates, calibration, and prototype refinement~\cite{nguyen2025adaptivecacheenhancementtesttime,liang2025advancing,zhang2024dual,zhu2025dynamic,zhai2025mitigating,chen2025multi,chen2025paf,guanstatistics}. However, they require sufficient target samples to construct reliable prototypes, creating a cold-start problem when the cache is initialized or reset. Target videos may also omit some expression classes, producing incomplete caches. Retaining source prototypes addresses missing classes~\cite{sharafi2026test}, but requires additional storage and is not personalized to the current subject. Moreover, correlated temporal windows may fill the cache with redundant embeddings, while noisy pseudo-labels propagate errors. These limitations motivate generating diverse class-wise prototypes personalized to each target video.

\section{TTA through Cache Personalization}

The core of \ours is a personalized sampled cache that generates diverse class-wise prototypes from each incoming target video. This cache is rebuilt online for every video, providing immediate subject-adaptive representations without requiring the accumulation of large target data or the retention of source prototypes. \ours further maintains two target caches across videos from the same subject: a positive cache of reliable target embeddings and a negative cache of uncertain embeddings. These target caches are reset when a new subject is encountered. For prediction, the current video embedding is compared with all cached representations, and the resulting similarities are converted into cache scores and combined with the CLIP scores. Adaptive entropy and diversity gates control updates to the target caches.

\begin{figure*}[t!]
  \centering
  \includegraphics[width=0.9\linewidth]{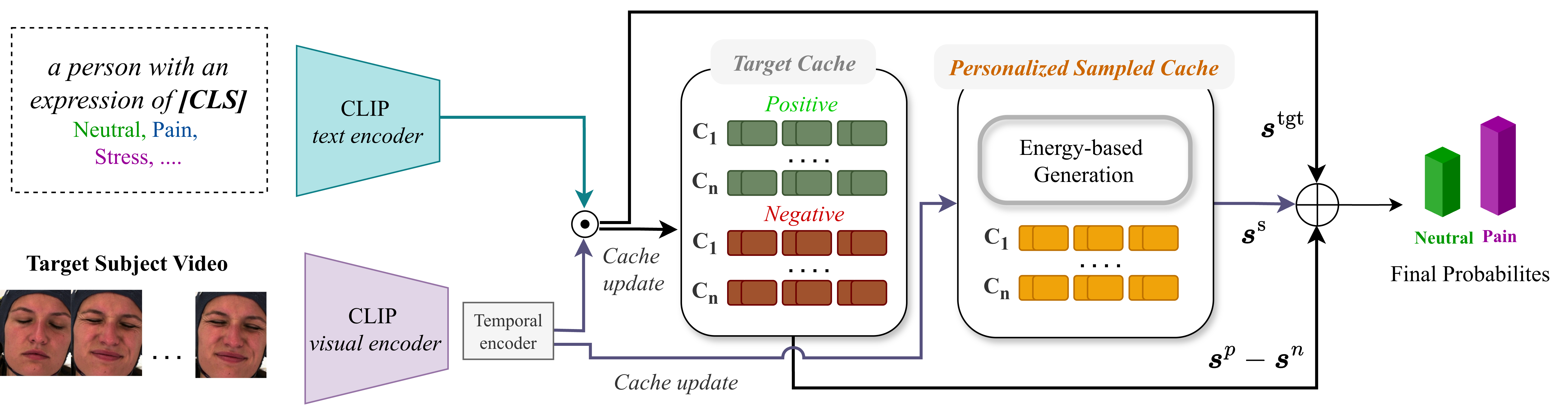}
\caption{Overview of \ours for online TTA. Frozen CLIP encoders and a temporal encoder produce the current video representation and CLIP class scores. An energy-based model (EBM) samples class-conditioned embeddings initialized from the target video to construct a personalized sampled cache covering all classes. In parallel, adaptive entropy and diversity gates control updates to the positive and negative target caches. Final predictions combine the current CLIP prediction with the caches.}
  \label{fig:Main_fig}
\end{figure*}

Figure~\ref{fig:Main_fig} summarizes the proposed method. Let $\mathrm{tgt}$ denote a test target subject and let $\mathcal{X}^{\mathrm{tgt}}$
represent an unlabeled video sequence of length $T$ from the target subject tgt. A frozen CLIP image encoder
$E_{\mathrm{I}}: \mathbb{R}^{H\times W\times 3}
\rightarrow \mathbb{R}^{d}$ maps each frame to an embedding
$\mathbf{v}^{\mathrm{tgt}}$. A temporal Transformer encoder
$E_{\mathrm{V}}:(\mathbb{R}^{d})^{L}\rightarrow\mathbb{R}^{d}$
models the dependencies among $L$ consecutive frame embeddings and produces the video-window representation
$\mathbf{z}^{\mathrm{tgt}}$. The temporal encoder is trained on the source data and remains fixed during test-time adaptation.
Let $\mathcal{C}$ denote the set of expression classes, and let $p_c$ be the textual prompt associated with class $c\in\mathcal{C}$. The frozen CLIP text encoder $E_{\mathrm{T}}$ produces the normalized class embedding $\mathbf{e}_c$. Given the current video representation $\mathbf{z}^{\mathrm{tgt}}$, the
CLIP class score for class $c$ is computed as
$s^{\mathrm{tgt}}(c)
=
\eta\,\cos\!\left(
\mathbf{z}^{\mathrm{tgt}},
\mathbf{e}_c
\right)$,
where $\eta>0$ is the fixed CLIP logit scale.

\subsection{Personalized Sampled Cache}
\label{sec:ebm_cache}

For each target video, a personalized sampled cache is constructed by drawing class-conditioned embeddings initialized from its normalized video representation $\mathbf{z}^{\mathrm{tgt}}$. The frozen recognition model, parameterized by $\theta$ and trained on the source data, defines a learned class-conditional energy through its class compatibility score ~\cite{duvenaud2020your}: \begin{equation} E_{\theta}(\mathbf{z},c) = -\mathbf{z}^{\top}\mathbf{e}_{c}, \label{eq:class_energy} \end{equation}
where $\mathbf{e}_{c}$ is the normalized class embedding produced by the frozen model, and lower energy indicates greater compatibility between representation $\mathbf{z}$ and class $c$. 

For every class $c\in\mathcal{C}$, a separate stochastic gradient Langevin dynamics (SGLD)~\cite{welling2011bayesian} chain is initialized from the current target representation: ${\mathbf{z}}_{0} = \mathbf{z}^{\mathrm{tgt}}$.
The chain is subsequently refined as \begin{equation} \widetilde{\mathbf{z}}_{k+1} = \widetilde{\mathbf{z}}_{k} - \frac{\alpha}{2} \nabla_{\widetilde{\mathbf{z}}_{k}} E_{\theta}(\widetilde{\mathbf{z}}_{k},c) + \sqrt{\alpha}\,\sigma\boldsymbol{\epsilon}_{k}, \qquad \boldsymbol{\epsilon}_{k} \sim \mathcal{N}(\mathbf{0},\mathbf{I}), \label{eq:sgld_update} \end{equation}
where $\alpha$ is the SGLD step size, $\sigma$ controls the noise magnitude, and $\boldsymbol{\epsilon}_{k}$ is independently sampled at each iteration from the standard Gaussian distribution $\mathcal{N}(\mathbf{0},\mathbf{I})$. The sampled embedding is $\ell_2$-normalized after each refinement step to remain in the normalized representation space of the recognition model.

Once the sampled embedding is predicted as the conditioning class $c$, the chain is stopped to encourage diversity without introducing an additional confidence threshold. Specifically, sampling stops when 
$ \arg\max_{c'\in\mathcal{C}} \widetilde{\mathbf{z}}_{k}^{\top}\mathbf{e}_{c'} = c$ or when the maximum number of refinement steps is reached. The final sampled embedding is then $\ell_2$-normalized before being stored in the cache. The recognition model remains fixed throughout this process, and gradients are computed only with respect to the sampled embedding.

After sampling, the personalized sampled cache is defined as
$\mathcal{K}^{s}
=
\{\widetilde{\mathbf{z}}_{c,j}
\mid c\in\mathcal{C},\,j=1,\ldots,m_s\}$, where $m_s$ embeddings are sampled for every class, and the class index $c$ indicates the corresponding conditioning class. Multiple chains can be sampled in parallel, including several samples from the same class. Since only the embedding is updated and the chains are stopped early, sampling requires few iterations and introduces limited computational overhead.

Initializing each chain from the current target representation preserves
video-specific characteristics, while independently sampled Gaussian noise
produces diverse embeddings for each class. A new $\mathcal{K}^{s}$ is
constructed for every target video and cleared before processing the next
one. Because the cache contains samples for all classes from the beginning,
it also alleviates the cold-start problem and provides evidence for classes
that have not yet been observed in the target caches.

\subsection{Target Caches}
\label{sec:tar_cache}

In addition to $\mathcal{K}^{s}$, \ours maintains a positive target cache $\mathcal{K}^{p}$ and a negative target cache $\mathcal{K}^{n}$. Both caches are updated sequentially as the temporal representations of the current video are processed. Their capacities are bounded per class to maintain constant memory usage. Retrieval is completed before the current representation is considered for insertion, ensuring that the current sample does not contribute to its own refined prediction.

\noindent\textbf{Positive cache.}
The positive cache is defined as
$\mathcal{K}^{p}
=
\{\mathbf{z}^{p}_{c,j}
\mid c\in\mathcal{C},\,j=1,\ldots,m_p\}$,
where $\mathbf{z}^{p}_{c,j}$ is the $j$-th accepted target embedding with
pseudo-label $c$, and $m_p$ is the current number of positive entries
for that class. Each class partition has a fixed capacity $m_{p}$. The
predictive entropy of each entry is retained only as metadata for
reliability-based cache replacement.

\noindent\textbf{Negative cache.}
The negative cache is defined as
$\mathcal{K}^{n}
=
\{\mathbf{z}^{n}_{c,j}
\mid c\in\mathcal{C},\,j=1,\ldots,m_n\}$,
where $\mathbf{z}^{n}_{c,j}$ is the $j$-th uncertain target embedding for
which class $c$ is the least likely prediction, and $m_n$ is the
current number of negative entries for that class. Each class partition has
a fixed capacity $m^{-}$, and the stored entropy is used only for cache
replacement.

\subsection{Online TTA per User Video}
\label{sec:online_tta}

For each target video, \ours processes the temporal representations sequentially. The current representation $\mathbf{z}^{\mathrm{tgt}}$ produces CLIP logits and retrieves evidence from $\mathcal{K}^{s}$, $\mathcal{K}^{p}$, and $\mathcal{K}^{n}$. The corresponding cache logits are combined with the current CLIP logits to obtain refined class scores. Cache updates are then controlled by an adaptive entropy gate for reliability and a diversity gate for representational coverage.


\noindent\textbf{Adaptive entropy gate.}
The adaptive entropy gate controls whether the current target representation updates the dynamic caches. Predictive entropy is computed from the refined class probabilities, where lower entropy indicates higher confidence. During the first five temporal representations of each video, fixed thresholds of $0.5$ and $0.8$ are used for the positive and negative caches, respectively. Afterwards, the thresholds are adapted using the running mean $\mu$ and standard deviation $\sigma$ of the predictive entropy values observed within the current video, as $\tau_p=\mu-\sigma$ and $\tau_n=\mu+\sigma$. Samples with entropy below $\tau_p$ are assigned their most probable class and considered for the positive cache, while samples with entropy between $\tau_p$ and $\tau_n$ are assigned their least probable class and considered for the negative cache. Samples above $\tau_n$ are rejected. The running entropy statistics are reset at the beginning of each video.

\noindent\textbf{Diversity gate.}
While entropy measures candidate reliability, it does not indicate whether the candidate adds new information to the cache. Diversity is measured as the mean feature-wise variance of the embeddings stored in the corresponding class partition of $\mathcal{K}^{p}$ or $\mathcal{K}^{n}$. For a candidate assigned to class $c$, this variance is computed before and after temporarily adding the candidate, and the diversity gain is defined as $\Delta D=D_{\mathrm{after}}-D_{\mathrm{before}}$. A positive gain indicates that the candidate increases the spread of the class representations, whereas a non-positive gain indicates limited or redundant information. The candidate is retained when $\Delta D>0$.

\subsection{Cache Retrieval and Final Prediction}
\label{sec:logit_fusion}

For the current representation $\mathbf{z}^{\mathrm{tgt}}$, cosine similarities are computed with all embeddings stored in the sampled, positive, and negative caches. For each class, the similarity-weighted contributions of all cache entries associated with that class are aggregated
to produce one sampled-cache score $\boldsymbol{s}^{s}$, positive-cache score $\boldsymbol{s}^{p}$, and negative-cache score
$\boldsymbol{s}^{n}$. Thus, no single prototype is selected; all available prototypes contribute to the corresponding class score. The refined class scores are obtained as
\begin{equation}
\boldsymbol{s}^{\mathrm{fuse}}
=
\boldsymbol{s}^{\mathrm{tgt}}
+
\boldsymbol{s}^{s}
+
\boldsymbol{s}^{p}
-
\boldsymbol{s}^{n},
\label{eq:logit_fusion}
\end{equation}
The final prediction is obtained by selecting the class with the highest fused score,
$\widehat{y}=\arg\max_{c\in\mathcal{C}}s^{\mathrm{fuse}}(c)$.

\begin{table*}[t!]
\centering

\scriptsize
\resizebox{0.75\linewidth}{!}{%
\begin{tabular}{l|l|cc|cc|cc}
\toprule
\multicolumn{1}{l|}{\multirow{2}{*}{\textbf{Protocol}}} &
\multicolumn{1}{l|}{\multirow{2}{*}{\textbf{Method}}} &
\multicolumn{2}{c|}{\textbf{BioVid}} &
\multicolumn{2}{c|}{\textbf{StressID}} &
\multicolumn{2}{c}{\textbf{BAH}} \\
\cmidrule(lr){3-4} \cmidrule(lr){5-6} \cmidrule(lr){7-8}
& & \war $\uparrow$ & \fonescore $\uparrow$ & \war $\uparrow$ & \fonescore $\uparrow$ & \war $\uparrow$ & \fonescore $\uparrow$ \\
\midrule

ZS & CLIP-ViT-B/32~\cite{radford2021learning}
{\fontsize{6}{12}\selectfont (ICML'21)} & 50.0 & 33.3 & 60.4 & 34.8 & 39.5 & 28.1 \\
\midrule

 
& CLIP-ViT-B/32~\cite{radford2021learning}
{\fontsize{6}{12}\selectfont (ICML'21)} & 69.7 & 66.6 & 67.0 & 44.5 & 60.4 & 39.8 \\

& EmoCLIP~\cite{foteinopoulou2024emoclip}{\fontsize{6}{12} \selectfont (FG'24)} & 67.7 & 63.4 & 63.5 & 35.9 & 56.2 & 36.5 \\

& X-CLIP~\cite{ni2022expanding}{\fontsize{6}{12} \selectfont (ECCV'22)} & 70.9 & 57.9 & 62.3 & 41.3 & 63.0 & 39.2 \\

\multirow{-4}{*}{FT} & Exp-CLIP~\cite{zhao2025enhancing} {\fontsize{6}{12} \selectfont (WACV'25)}& 70.2 & 66.7 & 63.1 & 44.5 & 62.2 & 38.5 \\
\midrule


& TPT~\cite{shu2022test}{\fontsize{6}{12} \selectfont (NeurIPS'22)} & 71.1 & 67.5 & 70.9 & 57.9 & 65.6 & 39.7 \\

& TDA~\cite{karmanov2024efficient}{\fontsize{6}{12} \selectfont (CVPR'24)} & 71.4 & 68.2 & 69.7 & 49.9 & 65.2 & 39.9 \\

& DPE~\cite{zhang2024dual}{\fontsize{6}{12} \selectfont (NeurIPS'24)} & 73.1 & 69.6 & 71.3 & 54.2 & 66.7 & 39.4 \\

& PromptAlign~\cite{abdul2023align}{\fontsize{6}{12} \selectfont (NeurIPS'23)} & 75.3 & 71.6 & 74.6 & 53.2 & 67.1 & 39.7 \\

& ReTA~\cite{liang2025advancing}{\fontsize{6}{12} \selectfont (ACMMM'25)}& 75.1 & 71.3 & 71.8 & 52.8 & 67.6 & 39.8 \\

& T3AL~\cite{liberatori2024test}{\fontsize{6}{12} \selectfont (CVPR'24)} & 76.1 & 72.9 & 75.9 & 59.4 & 67.9 & 40.7 \\

\rowcolor{blueTTA}
\cellcolor{white}\multirow{-7}{*}{FT\,+\,TTA} & {\textbf{\ours (ours)}} & \textbf{81.0} & \textbf{78.2} & \textbf{81.2} & \textbf{67.0} & \textbf{68.9} & \textbf{41.0} \\
\bottomrule
\end{tabular}
}
\caption{Performance (\fonescore score and \war) of zero-shot (ZS), fine-tuning (FT), and TTA on CLIP models for FER. Results are averaged over 10 target subjects for each dataset. CLIP-ViT-B/32\textsuperscript{$\dagger$} denotes full CLIP fine-tuning.}

\label{tab:source_vs_tta_results}
\end{table*}

\begin{table*}[t!]
\centering
\scriptsize

\resizebox{0.9\linewidth}{!}{%
\begin{tabular}{l|cccccccccc|c}
\toprule
\textbf{Method} & Sub-1 & Sub-2 & Sub-3 & Sub-4 & Sub-5 & Sub-6 & Sub-7 & Sub-8 & Sub-9 & Sub-10 & Avg. \\
\midrule
TPT~\cite{shu2022test}          & 92.0 & 51.5 & 40.0 & 49.6 & 79.0 & 87.0 & 79.5 & 43.4 & 99.0 & 54.0 & 67.5 \\
TDA~\cite{karmanov2024efficient}           & 94.9 & 50.3 & 43.5 & 48.0 & 80.8 & 86.2 & 80.2 & 43.5 & \textbf{100.0} & 55.0 & 68.2 \\
DPE~\cite{zhang2024dual}          & 90.4 & 51.5 & 43.5 & 73.1 & 72.0 & 83.6 & 79.8 & 48.5 & \textbf{100.0} & 54.4 & 69.6 \\
PromptAlign~\cite{abdul2023align}  & 94.2 & 60.1 & 42.5 & 61.5 & 83.5 & 90.5 & 80.2 & 43.5 & \textbf{100.0} & 60.0 & 71.6 \\
ReTA~\cite{liang2025advancing}         & 94.2 & 60.1 & 43.5 & 60.8 & 83.5 & 88.0 & 80.2 & 43.5 & \textbf{100.0} & 60.0 & 71.3 \\
T3AL~\cite{liberatori2024test}         & \textbf{95.8} & 62.0 & \textbf{43.9} & 61.5 & 83.9 & 93.9 & 83.0 & 43.5 & \textbf{100.0} & 62.0 & 72.9 \\

\rowcolor{blueTTA}
\textbf{\ours (ours)} & 94.4 & \textbf{66.0} & 33.3 & \textbf{73.0} & \textbf{99.5} & \textbf{97.1} & \textbf{89.0} & \textbf{47.0} & \textbf{100.0} & \textbf{83.3} & \textbf{78.2} \\

\bottomrule
\end{tabular}%
 }
\caption{\fonescore score per subject on the \biovid dataset for \ours and competing TTA methods. Best results are shown in \textbf{bold}.}
\label{tab:biovid_subject_full_metrics_f1}
\end{table*}

\begin{table*}[t!]
\centering
\scriptsize

\resizebox{0.9\linewidth}{!}{%
\begin{tabular}{l|cccccccccc|c}
\toprule
\textbf{Method} & Sub-1 & Sub-2 & Sub-3 & Sub-4 & Sub-5 & Sub-6 & Sub-7 & Sub-8 & Sub-9 & Sub-10 & Avg. \\
\midrule
TPT~\cite{shu2022test}          & 70.7 & 43.6 & 86.0 & \textbf{83.1} & 36.4 & 78.7 & 41.1 & 50.0 & 40.0 & 50.0 & 57.9 \\
TDA~\cite{karmanov2024efficient}          & 45.5 & 40.0 & 50.1 & 81.0 & 35.0 & 80.0 & 29.6 & 45.6 & 42.9 & 50.0 & 49.9 \\
DPE~\cite{zhang2024dual}          & 41.0 & 40.6 & 49.0 & 81.0 & 49.8 & 80.0 & 56.0 & 46.0 & 42.9 & 56.6 & 54.2 \\
PromptAlign~\cite{abdul2023align}  & 41.3 & 45.0 & 45.6 & 82.3 & 48.1 & 80.0 & 44.0 & 58.0 & 42.9 & 45.5 & 53.2 \\
ReTA~\cite{liang2025advancing}         & 43.2 & 41.0 & 53.0 & 80.9 & 36.1 & 80.0 & 39.1 & 53.3 & 42.9 & 59.3 & 52.8 \\
T3AL~\cite{liberatori2024test}         & 46.3 & 48.6 & 48.6 & 80.1 & \textbf{51.0} & 80.0 & 61.8 & 68.0 & 44.0 & 66.0 & 59.4 \\
\rowcolor{blueTTA}
\textbf{\ours (ours)} & \textbf{77.0} & \textbf{50.0} & \textbf{87.0} & 82.0 & 46.1 & \textbf{84.2} & \textbf{62.0} & \textbf{68.5} & \textbf{46.2} & \textbf{67.1} & \textbf{67.0} \\

\bottomrule
\end{tabular}%
 }
\caption{\fonescore score per subject on \stressid for \ours and other TTA methods.}
\label{tab:stressid_subject_full_metrics_f1}
\end{table*}
%
%
\begin{table*}[t!]
\centering
\scriptsize
\resizebox{0.9\linewidth}{!}{%
\begin{tabular}{l|cccccccccc|c}
\toprule
\textbf{Method} & Sub-1 & Sub-2 & Sub-3 & Sub-4 & Sub-5 & Sub-6 & Sub-7 & Sub-8 & Sub-9 & Sub-10 & Avg. \\
\midrule
TPT~\cite{shu2022test} & 40.6 & 35.7 & 43.7 & 45.9 & 45.6 & 43.1 & 35.0 & 34.7 & 40.7 & 32.9 & 39.7 \\
TDA~\cite{karmanov2024efficient} & 40.0 & 39.7 & 44.4 & 46.0 & 42.0 & 42.6 & 35.0 & 35.5 & 40.7 & 33.2 & 39.9 \\
DPE~\cite{zhang2024dual} & 34.8 & 37.1 & 44.2 & 46.0 & 45.3 & 43.0 & 37.2 & 32.5 & \textbf{41.0} & 33.2 & 39.4 \\
PromptAlign~\cite{abdul2023align} & 25.0 & 37.0 & 46.9 & \textbf{51.0} & 46.1 & 43.5 & 41.0 & 32.5 & \textbf{41.0} & 33.2 & 39.7 \\
ReTA~\cite{liang2025advancing} & 25.0 & 36.9 & 41.9 & \textbf{51.0} & 46.1 & 51.2 & 40.0 & 32.5 & \textbf{41.0} & 33.2 & 39.8 \\
T3AL~\cite{liberatori2024test} & 41.6 & 37.0 & \textbf{44.7} & 46.0 & 45.2 & 43.0 & 38.0 & \textbf{36.2} & \textbf{41.0} & 35.0 & 40.7 \\
\rowcolor{blueTTA}
\textbf{\ours (ours)} & \textbf{41.8} & \textbf{39.9} & \textbf{44.7} & 45.9 & \textbf{46.2} & \textbf{43.6} & \textbf{41.0} & 32.0 & 40.4 & \textbf{35.2} & \textbf{41.0} \\
\bottomrule
\end{tabular}%
 }
\caption{\fonescore score per subject on \bah for \ours and other TTA methods.}
\label{tab:bah_subject_full_metrics_f1}
\end{table*}

\section{Results and Discussion}

\subsection{Experimental Methodology}

\noindent\textbf{Datasets.}  
Three challenging public video datasets are used to evaluate the proposed approach: \biovid~\cite{walter2013biovid}, \stressid~\cite{chaptoukaev2023stressid}, and \bah~\cite{gonzalez2025bah}. In line with previous studies~\cite{zeeshan2024subject, sharafi2025disentangled, zeeshan2025musaco, zeeshan2025progressive}, subject-independent (cross-subject) splits are employed to construct the source and target sets. \biovid Heat Pain Database (Part A) consists of controlled heat-pain recordings spanning several intensity levels plus a neutral condition, accompanied by synchronized facial video; the source set comprises 77 subjects, while the remaining 10 form the target set. \stressid provides facial recordings gathered during stress-inducing tasks along with corresponding self-reported affect labels, split into 44 source subjects and 10 target subjects. \bah is a large-scale, multimodal dataset for recognizing ambivalence and hesitancy, captured via webcam and annotated at the frame level; it is divided into 143 source subjects and 10 target subjects.

\noindent\textbf{Implementation Details.}
All experiments were conducted on a single NVIDIA A100 GPU with 48\,GB of memory, using CLIP ViT-B/32 as the backbone and a batch size of 16. For each target video, the class-conditional sampler generates $m_s{=}3$ embeddings per class using at most $20$ SGLD iterations, with a step size of $0.01$ and a noise scale of $0.1$. The positive and negative cache capacities are set to
$m_p{=}5$ and $m_n{=}4$ embeddings per class, respectively. The running entropy statistics are reset at the beginning of each video. During the first five temporal representations, a positive entropy threshold of $0.5$ and a negative entropy threshold of $0.8$ are used for warm-up. The thresholds are then adapted using the entropy statistics accumulated within the same video, with
both adaptation coefficients set to $1$. The diversity gate retains a candidate only when it produces a positive variance gain, $\Delta D>0$,
and negative cache scores are fused using unit weights. The same configuration is used for all datasets.

\begin{table}[b!]
\centering
\scriptsize
\resizebox{0.7\linewidth}{!}{%
\begin{tabular}{l|ccr}
\toprule
\textbf{Method} & \textbf{Time (ms)} & \textbf{Memory (MB)} & \multicolumn{1}{r}{\textbf{\war (\%)}}\\
\midrule
TPT & 771.2 & 2800 & 71.1\\
PromptAlign & 900.0 & 3100 & 75.3\\
TDA &  \textbf{97.8} & 2124 & 71.4\\
DPE & 263.9 & 1700 & 73.1\\
ReTA & 174.0 & 2390 & 75.1\\
T3AL & 520.0 & 2200 & 76.1\\
\rowcolor{pinkTTA}
\textbf{\ours} & 220.0 & \textbf{612.0} & \textbf{81.0}\\
\bottomrule
\end{tabular}
}
\caption{Complexity of TTA methods. Run time per batch (B=16), memory, and average \war for the \biovid dataset.}
\label{tab:runtime_biovid}
\end{table}

Baseline comparisons cover both CLIP-based FER approaches and TTA methods. For CLIP-based FER, \ours is benchmarked against the frozen CLIP-ViT-B/32 backbone in a zero-shot setting, CLIP-ViT-B/32\textsuperscript{ †}~\cite{foteinopoulou2024emoclip} (CLIP fine-tuned on the source domain), and EmoCLIP~\cite{foteinopoulou2024emoclip} (which trains only an adapter). Further baselines include X-CLIP~\cite{ni2022expanding} and Exp-CLIP~\cite{zhao2025enhancing}, which use trainable projection modules; Exp-CLIP additionally uses LLM-generated expression descriptions. For TTA, \ours is compared with methods developed for image recognition—namely TPT~\cite{shu2022test}, TDA~\cite{karmanov2024efficient}, DPE~\cite{zhang2024dual}, PromptAlign~\cite{abdul2023align}, and ReTA~\cite{liang2025advancing}—as well as an adapted version of T3AL~\cite{liberatori2024test}, a video-based action recognition approach. All baselines use the same CLIP backbone, prompts, preprocessing, and experimental protocol. Consistent with prior CLIP-based FER research~\cite{zhao2025enhancing}, the prompt template "\textit{a person with an expression of [CLS]}" is applied across all TTA baselines and \ours. Further details on baseline and prompt selection can be found in supplementary materials.

\begin{figure}[t!]
    \centering
    \includegraphics[
        width=0.75\linewidth,
        height=0.45\linewidth,
        keepaspectratio
    ]{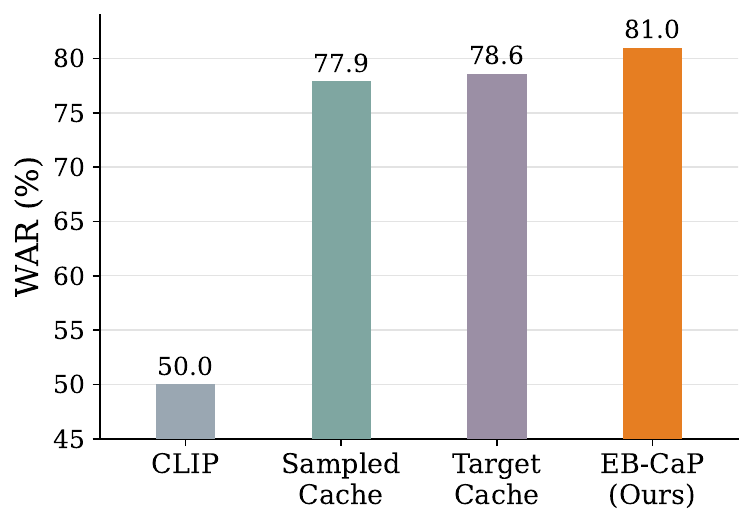}
    \caption{
    Cache design on \biovid~data. Results compare the CLIP
    baseline, personalized sampled cache, target cache
    (positive and negative), and complete \ours method in terms of
    \war.
    }
    \label{fig:cache_design}
\end{figure}

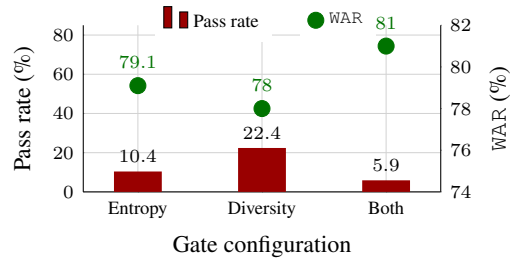
\begin{figure}[t!]
\centering
\begin{tikzpicture}

\def\EntropyPass{10.4}
\def\DiversityPass{22.4}
\def\BothPass{5.9}

\def\EntropyWAR{79.1}
\def\DiversityWAR{78.0}
\def\BothWAR{81.0}

\begin{axis}[
    width=0.75\linewidth,
    height=0.45\linewidth,
    ybar,
    bar width=18pt,
    ymin=0,
    ymax=85,
    ylabel={Pass rate (\%)},
    xlabel={Gate configuration},
    symbolic x coords={Entropy,Diversity,Both},
    xtick=data,
    tick label style={font=\scriptsize},
    label style={font=\small},
    axis y line*=left,
    axis x line*=bottom,
    enlarge x limits=0.22,
    grid=major,
    major grid style={
        line width=0.2pt,
        draw=gray!35
    },
    nodes near coords,
    nodes near coords style={
        font=\scriptsize,
        anchor=south,
        /pgf/number format/fixed,
        /pgf/number format/precision=1
    },
    legend style={
        font=\scriptsize,
        at={(0.20,1.15)},
        anchor=north west,
        draw=none
    },
    major tick length=0pt,
]

\addplot[
    draw=none,
    fill=red!60!black
] coordinates {
    (Entropy,\EntropyPass)
    (Diversity,\DiversityPass)
    (Both,\BothPass)
};

\addlegendentry{Pass rate}

\end{axis}

\begin{axis}[
    width=0.75\linewidth,
    height=0.45\linewidth,
    ymin=74,
    ymax=82,
    ylabel={\war\ (\%)},
    symbolic x coords={Entropy,Diversity,Both},
    xtick=\empty,
    axis y line*=right,
    axis x line=none,
    tick label style={font=\scriptsize},
    label style={font=\small},
    enlarge x limits=0.22,
    nodes near coords,
    nodes near coords style={
        font=\scriptsize,
        anchor=south,
        yshift=3pt,
        /pgf/number format/fixed,
        /pgf/number format/precision=1
    },
    legend style={
        font=\scriptsize,
        at={(0.82,1.14)},
        anchor=north east,
        draw=none
    },
    major tick length=0pt,
]

\addplot[
    only marks,
    mark=*,
    mark size=3pt,
    green!45!black
] coordinates {
    (Entropy,\EntropyWAR)
    (Diversity,\DiversityWAR)
    (Both,\BothWAR)
};

\addlegendentry{\war}

\end{axis}

\end{tikzpicture}

\caption{
Adaptive entropy and diversity gates on \biovid. Bars
report the percentage of candidates admitted by each gate configuration,
while points show the corresponding \war. ``Both'' indicates that both
gates are jointly applied.
}
\label{fig:gate_analysis}
\end{figure}

\subsection{Comparison with State-of-the-Art Methods}

Table~\ref{tab:source_vs_tta_results} compares zero-shot CLIP, fine-tuned FER models, and recent TTA approaches on \biovid, \stressid, and \bah. Fine-tuning and test-time adaptation generally improve over zero-shot inference, confirming that a fixed CLIP representation is insufficient to address the subject shift. Among the compared methods, \ours achieves the best \war and \fonescore on all three datasets. The improvements
are particularly pronounced on \biovid and \stressid, where \ours provides clear gains over the strongest competing TTA method in both metrics. On \bah, the improvement is smaller, indicating that the remaining recognition errors are more difficult to resolve through test-time personalization.

The subject-wise results in
Tables~\ref{tab:biovid_subject_full_metrics_f1}--\ref{tab:bah_subject_full_metrics_f1}
show that only a few favorable subjects do not cause the improvements. On \biovid and \stressid, \ours obtains the strongest performance for most target subjects and substantially improves the average \fonescore, although a small number of subjects remain challenging. The results on \bah are more mixed, with smaller differences among the methods and varying rankings across subjects. This subject-level variability highlights the importance of evaluating personalization methods beyond dataset-level averages. Conditioned sampled embeddings with reliable positive and complementary negative target caches
provide more robust adaptation than relying only on model optimization or observed target samples. Moreover, these improvements are obtained without updating the recognition model during inference. A paired two-sided Wilcoxon signed-rank test over the subject-wise \war results on \biovid confirms a statistically significant improvement over T3AL ($p=0.039<0.05$). Complete subject-wise \war results and additional CLIP-based FER comparisons are provided in the supplementary materials.

\noindent\textbf{Computation Complexity}. Table~\ref{tab:runtime_biovid} further examines the accuracy--efficiency trade-off. Although TDA achieves the lowest runtime, its recognition
performance remains substantially below that of \ours. In contrast, \ours achieves the highest \war while requiring the least GPU memory among all compared methods. Its runtime remains lower than optimization-intensive approaches such as TPT, PromptAlign, and T3AL, showing that the additional class-conditioned sampling introduces moderate computational cost without the memory overhead of test-time model updates. Overall, \ours provides the strongest balance between adaptation performance and resource efficiency.

\begin{table}[b!]
\centering
\resizebox{0.7\linewidth}{!}{%
\begin{tabular}{lc}
\toprule

Method & \war (\%) \\[-0.6ex]
\midrule
GMM-based                    & 74.0 \\
Classifier-based              & 77.6 \\
Generator-based                & 79.1 \\
Prototype$^{*}$~\cite{sharafi2026test}   & 81.5 \\
EEnergy-based with noise initialization & 78.2 \\ Energy-based with target initialization (ours) & 81.0 \\
\bottomrule
\end{tabular}
}

\caption{Comparison of personalized cache construction strategies on \biovid. Prototype-based cache construction requires source-derived class prototypes, whereas \ours constructs the cache through target-initialized energy-based sampling without source information. $^{*}$Requires source-derived class prototypes.}
\label{tab:source_cache_ablation}

\end{table}

\subsection{Ablation Studies}

\noindent\textbf{Ablation on Cache Design.}
Figure~\ref{fig:cache_design} evaluates the contribution of the proposed cache components on \biovid. The CLIP baseline achieves a \war of $50.0\%$, while using only the personalized sampled cache increases performance to $77.9\%$. The target cache alone reaches $78.6\%$, showing the benefit of subject-specific test-time information. Combining the sampled and target caches in \ours yields the best result of $81.0\%$, outperforming the individual caches by $3.1$ and $2.4$ percentage points, respectively. These results indicate that the generated class-specific representations and the dynamically collected target embeddings provide complementary information for subject personalization.

\noindent\textbf{Ablation on Cache-Update Gating.}
Figure~\ref{fig:gate_analysis} compares the adaptive entropy gate, the diversity gate, and their joint use on \biovid. The entropy-only and diversity-only configurations achieve \war scores of 79.1\% and 78.0\%, with pass rates of 10.4\% and 22.4\%, respectively. Applying both gates reduces the pass rate to 5.9\% while improving \war to 81.0\%. These results indicate that jointly enforcing reliability and diversity produces a more selective cache and yields the best recognition performance.

\begin{figure}[t!]
    \centering
    \includegraphics[
        width=0.85\linewidth,
        height=0.55\linewidth,
        keepaspectratio
    ]{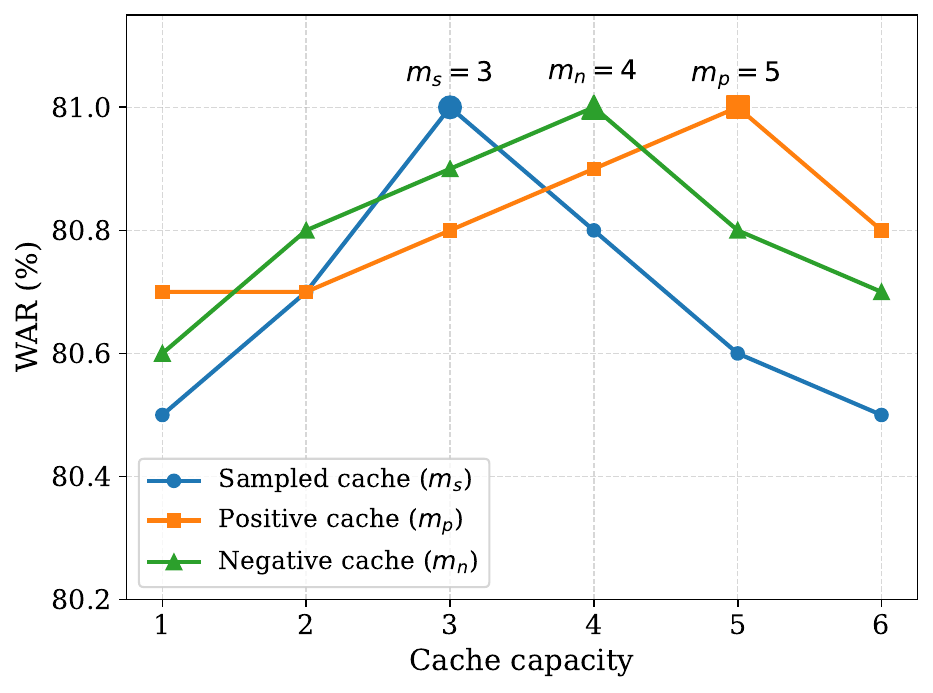}
    \caption{Sensitivity analysis of the sampled, positive, and negative
    cache capacities on \biovid.}
    \label{fig:cache_capacity}
\end{figure}

\noindent\textbf{Cache construction.}
Table~\ref{tab:source_cache_ablation} compares different personalized cache construction strategies. The GMM-, classifier-, and generator-based variants rely on representations learned from labeled source data, while noise-initialized energy-based sampling generates class-conditioned embeddings without using the current target representation. The prototype-based strategy~\cite{sharafi2026test} is included as a reference but requires source-derived class prototypes at test time. In contrast, the proposed target-initialized energy-based strategy (ours) uses no source samples, labels, or prototypes and initializes each SGLD chain from the current unlabeled target video. Despite this restriction, \ours achieves competitive performance, showing that target-initialized sampling can construct effective class-specific caches without retaining source information.

\noindent\textbf{Cache capacity.}
Figure~\ref{fig:cache_capacity} studies the effect of the sampled, positive, and negative cache capacities in \ours. Each capacity is varied independently while the remaining two are fixed. Performance remains relatively stable across the evaluated settings, indicating limited sensitivity to moderate changes in capacity. The best configuration for \biovid is $m_s{=}3$, $m_p{=}5$, and $m_n{=}4$.

\section{Conclusion}
This paper introduced \ours, a personalized TTA method for fine-grained video FER. For each unlabeled target video, \ours constructs a personalized sampled cache by generating class-wise embeddings using SGLD that is initialized from the current video representation. This cache is complemented by positive and negative target caches that retain reliable and uncertain observations. Adaptive entropy and diversity gates control cache updates to reduce noisy pseudo-label accumulation and redundant representations. The resulting cache-derived class scores are fused with the current CLIP scores while the recognition model remains fixed. Experiments on three challenging video datasets including \biovid, \stressid, and \bah demonstrate that \ours consistently improves over existing fine-tuning and TTA methods without requiring source data, target labels, or test-time model updates. 

\noindent\textbf{Supplementary materials.}
Includes additional subject-wise \war results for TTA methods, baseline and implementation details, code, the online TTA per video algorithm, and extended ablations on prompt selection.

\bibliography{aaai2027}


\end{document}


\maketitle

\begin{center}
{\large\textbf{Table of Contents}}
\end{center}

{
\noindent
\textbf{1\quad Algorithmic Details}
\dotfill
\pageref{sec:algorithmic_details}
\par

\vspace{0.2cm}

\noindent
\textbf{2\quad Additional Experimental Results}
\dotfill
\pageref{sec:additional_results}
\par

\vspace{0.2cm}

\noindent
\hspace*{1.5em}
2.1\quad TTA Baseline Implementation Details
\dotfill
\pageref{subsec:tta_baselines}
\par

\vspace{0.2cm}

\noindent
\hspace*{1.5em}
2.2\quad Subject-Wise Results
\dotfill
\pageref{subsec:subject_results}
\par

\vspace{0.2cm}

\noindent
\hspace*{1.5em}
2.3\quad Additional Ablation Studies
\dotfill
\pageref{subsec:additional_ablations}
\par
}

\vspace{1em}

\section{Algorithmic Details}
\label{sec:algorithmic_details}

This section provides additional algorithmic details for the online test-time adaptation pipeline introduced in the main paper. Algorithm~\ref{alg:ebcap} summarizes the complete inference procedure of \ours for each target subject and video. The frozen CLIP image encoder $E_{\mathrm{I}}$, temporal encoder $E_{\mathrm{V}}$, and text encoder $E_{\mathrm{T}}$ are first used to extract the target video representation and class text embeddings. For each temporal representation, a personalized sampled cache is generated through class-conditioned SGLD initialized from the current target embedding. Evidence retrieved from this sampled cache and the positive and negative target caches is then combined with the current CLIP scores to obtain the refined prediction. Finally, adaptive entropy and diversity gates determine whether the current target representation is inserted into the positive or negative cache. The sampled cache is reconstructed for every target representation, whereas the target caches are maintained across videos of the same subject and reset when a new subject is encountered. This procedure enables fully online subject personalization without retaining source samples or prototypes and without updating the frozen recognition model at inference time.

\begin{algorithm}[t]
\caption{Online Test-Time Adaptation with \ours}
\label{alg:ebcap}
\small
\begin{algorithmic}[1]

\REQUIRE Target video stream, frozen encoders
$E_{\mathrm{I}}$, $E_{\mathrm{V}}$, and $E_{\mathrm{T}}$,
class prompts $\{p_c\}_{c\in\mathcal{C}}$

\ENSURE Refined predictions $\widehat{y}$

\STATE $\mathbf{e}_c \leftarrow
\operatorname{Normalize}(E_{\mathrm{T}}(p_c))$,
$\forall c\in\mathcal{C}$

\FOR{each target subject}
    \STATE Initialize
    $\mathcal{K}^{p},\mathcal{K}^{n}\leftarrow\emptyset$

    \FOR{each target video $\mathcal{X}^{\mathrm{tgt}}$}
        \STATE Reset video-level entropy statistics

        \FOR{each temporal window $\mathbf{X}^{\mathrm{tgt}}_r$}

            \STATE $\mathbf{z}^{\mathrm{tgt}}\leftarrow
            \operatorname{Normalize}\!\left(
            E_{\mathrm{V}}\!\left(
            E_{\mathrm{I}}(\mathbf{X}^{\mathrm{tgt}}_r)
            \right)\right)$

            \STATE Compute CLIP scores
            $\boldsymbol{s}^{\mathrm{tgt}}$ using
            $\mathbf{z}^{\mathrm{tgt}}$ and
            $\{\mathbf{e}_c\}_{c\in\mathcal{C}}$

            \STATE Generate $\mathcal{K}^{s}$ using
            class-conditioned SGLD initialized from
            $\mathbf{z}^{\mathrm{tgt}}$

            \STATE Compute cache scores
            $\boldsymbol{s}^{s}$,
            $\boldsymbol{s}^{p}$, and
            $\boldsymbol{s}^{n}$

            \STATE $\boldsymbol{s}^{\mathrm{fuse}}
            \leftarrow
            \boldsymbol{s}^{\mathrm{tgt}}
            +\boldsymbol{s}^{s}
            +\boldsymbol{s}^{p}
            -\boldsymbol{s}^{n}$

            \STATE $\widehat{y}\leftarrow
            \arg\max_{c\in\mathcal{C}}
            s^{\mathrm{fuse}}(c)$

            \STATE Compute entropy $H$ and update
            thresholds $\tau_p$ and $\tau_n$

            \IF{$H<\tau_p$}
                \STATE Update $\mathcal{K}^{p}$ with
                $\mathbf{z}^{\mathrm{tgt}}$ if $\Delta D>0$
            \ELSIF{$H\leq\tau_n$}
                \STATE Update $\mathcal{K}^{n}$ with
                $\mathbf{z}^{\mathrm{tgt}}$ if $\Delta D>0$
            \ENDIF

            \STATE Discard $\mathcal{K}^{s}$

        \ENDFOR
    \ENDFOR
\ENDFOR

\end{algorithmic}
\end{algorithm}

\section{Additional Experimental Results}
\label{sec:additional_results}


\begin{table*}[t]
\centering
\scriptsize
\resizebox{0.95\linewidth}{!}{%
\begin{tabular}{l|cccccccccc|c}
\toprule
\textbf{Method} & Sub-1 & Sub-2 & Sub-3 & Sub-4 & Sub-5 &
Sub-6 & Sub-7 & Sub-8 & Sub-9 & Sub-10 & Avg. \\
\midrule
TPT~\cite{shu2022test} 
& 91.9 & 53.0 & 50.0 & 61.5 & 79.0
& 88.9 & 76.0 & 50.0 & 98.9 & 62.0 & 71.1 \\
TDA~\cite{karmanov2024efficient} 
& 93.0 & 52.6 & 50.0 & 63.9 & 79.5
& 87.5 & 76.3 & 50.0 & 100.0 & 61.6 & 71.4 \\
DPE~\cite{zhang2024dual}
& 91.7 & 55.0 & 50.0 & 85.0 & 73.0
& 84.7 & 80.0 & 50.0 & 100.0 & 61.8 & 73.1 \\
PromptAlign~\cite{abdul2023align} 
& 93.8 & 65.2 & 50.0 & 70.0 & 85.0
& 92.0 & 80.0 & 50.0 & 100.0 & 67.6 & 75.3 \\
ReTA~\cite{liang2025advancing} 
& 93.8 & 65.2 & 50.0 & 69.5 & 85.0
& 90.2 & 80.0 & 50.0 & 100.0 & 67.6 & 75.1 \\
T3AL~\cite{liberatori2024test} 
& 94.0 & 66.5 & 50.0 & 70.0 & 85.2
& 94.2 & 83.9 & 50.0 & 100.0 & 67.6 & 76.1 \\
\rowcolor{blueTTA}
\textbf{\ours (ours)} 
& \textbf{94.8} & \textbf{71.9} & \textbf{50.0} & \textbf{95.4} &  \textbf{92.1}& \textbf{97.5} & \textbf{90.8} & \textbf{50.0} & \textbf{100.0} & \textbf{68.0} & \textbf{81.0} \\
\bottomrule
\end{tabular}%
}

\caption{\war per subject on the \biovid dataset for \ours and competing
TTA methods. Best results are shown in \textbf{bold}.}
\label{tab:biovid_subject_war}
\end{table*}


\begin{table*}[t]
\centering
\scriptsize

\resizebox{0.95\linewidth}{!}{%
\begin{tabular}{l|cccccccccc|c}
\toprule
\textbf{Method} & Sub-1 & Sub-2 & Sub-3 & Sub-4 & Sub-5 &
Sub-6 & Sub-7 & Sub-8 & Sub-9 & Sub-10 & Avg. \\
\midrule

TPT~\cite{shu2022test} 
& 74.7 & 51.6 & 90.0 & 88.0 & 52.2
& 77.5 & 66.6 & 63.4 & 54.0 & 91.9 & 70.9 \\
TDA~\cite{karmanov2024efficient} 
& 66.9 & 45.2 & 75.6 & 90.1 & 49.6
& 80.0 & 41.2 & 75.3 & 81.8 & 91.9 & 69.7 \\
DPE~\cite{zhang2024dual}
& 64.0 & 55.0 & 73.9 & 90.0 & 52.6
& 80.0 & 50.9 & 73.0 & 81.8 & 92.0 & 71.3 \\
PromptAlign~\cite{abdul2023align} 
& 69.9 & 61.6 & 78.6 & 90.9 & 51.2
& 80.0 & 60.4 & 81.3 & 81.8 & 90.9 & 74.6 \\
ReTA~\cite{liang2025advancing} 
& 64.5 & 50.5 & 82.0 & 90.0 & 46.8
& 80.0 & 49.3 & 80.0 & 81.8 & 93.7 & 71.8 \\
T3AL~\cite{liberatori2024test} 
& 71.3 & 62.7 & 80.3 & 90.2 & 54.0
& 80.0 & 68.0 & 77.0 & 82.0 & 93.9 & 75.9 \\

\rowcolor{blueTTA}
\textbf{\ours (ours)} 
& \textbf{81.1} & \textbf{66.9} & \textbf{91.2} & \textbf{89.0} & \textbf{69.9} & \textbf{81.2} & \textbf{69.5} & \textbf{83.8} & \textbf{84.6} & \textbf{95.7} & \textbf{81.2} \\

\bottomrule
\end{tabular}%
}

\caption{\war per subject on \stressid for \ours and other TTA methods.}
\label{tab:stressid_subject_war}
\end{table*}


\begin{table*}[t]
\centering
\scriptsize
\resizebox{0.95\linewidth}{!}{%
\begin{tabular}{l|cccccccccc|c}
\toprule
\textbf{Method} & Sub-1 & Sub-2 & Sub-3 & Sub-4 & Sub-5 &
Sub-6 & Sub-7 & Sub-8 & Sub-9 & Sub-10 & Avg. \\
\midrule
TPT~\cite{shu2022test} 
& 62.0 & 53.7 & 73.9 & 80.0 & 81.2
& 71.0 & 58.0 & 55.0 & 69.0 & 53.0 & 65.6 \\
TDA~\cite{karmanov2024efficient} 
& 58.0 & 53.7 & 75.0 & 80.3 & 79.0
& 71.6 & 58.0 & 55.9 & 69.0 & 51.7 & 65.2 \\
DPE~\cite{zhang2024dual}
& 58.6 & 59.6 & 79.6 & 80.5 & 83.4
& 67.0 & 59.0 & 54.0 & 71.9 & 54.0 & 66.7 \\
PromptAlign~\cite{abdul2023align} 
& 59.1 & 59.1 & 72.0 & 84.5 & 84.1
& 72.4 & 58.5 & 55.9 & 71.5 & \textbf{54.9} & 67.1 \\
ReTA~\cite{liang2025advancing} 
& 61.4 & 60.0 & 72.0 & 84.0 & 84.1
& 72.4 & 58.5 & 55.9 & \textbf{72.0} & 56.0 & 67.6 \\
T3AL~\cite{liberatori2024test} 
& 61.0 & 58.9 & 75.0 & \textbf{85.6} & 83.6
& \textbf{74.7} & 60.0 & 56.0 & 71.0 & 53.6 & 67.9 \\
\rowcolor{blueTTA}
\textbf{\ours (ours)} 
& \textbf{62.1} & \textbf{60.4} & \textbf{79.9} & \textbf{85.6} & \textbf{84.5} & \textbf{74.7} & \textbf{61.1} & \textbf{56.5} & \textbf{72.0} & 53.0 & \textbf{68.9} \\
\bottomrule
\end{tabular}%
}
\caption{\war per subject on \bah for \ours and other TTA methods.}
\label{tab:bah_subject_war}
\end{table*}

\subsection{TTA Baseline Implementation Details}
\label{subsec:tta_baselines}

Several CLIP-based test-time adaptation methods are included as baselines. \emph{TPT}~\cite{shu2022test} performs test-time prompt tuning by optimizing the learnable text prompt tokens through entropy minimization over multiple strongly augmented views, while leaving the CLIP image and text encoders unchanged. \emph{TDA}~\cite{karmanov2024efficient} avoids test-time optimization and instead maintains an online feature--label memory containing both positive and negative pseudo-label information. Predictions retrieved from this dynamic cache are combined with the original CLIP zero-shot scores. \emph{DPE}~\cite{zhang2024dual} jointly adapts visual and textual class prototypes during inference, augments them with lightweight instance-dependent residuals, and imposes cross-modal consistency to preserve agreement between the two prototype spaces under target-domain shifts. \emph{PromptAlign}~\cite{abdul2023align} updates the textual prompts by matching test-time feature statistics to statistics estimated from the source domain, providing an additional alignment objective beyond entropy minimization. \emph{ReTA}~\cite{liang2025advancing} improves cache reliability through consistency-aware entropy reweighting for selective memory updates and diversity-driven distribution calibration, where each class is represented by a Gaussian distribution over dynamically evolving text embeddings. Finally, \emph{T3AL}~\cite{liberatori2024test}, originally designed for zero-shot temporal action localization, is modified for video classification. The adapted implementation derives video-level pseudo-labels from aggregated frame representations, refines frame-level predictions through self-supervision, and uses textual guidance to suppress irrelevant temporal regions. Its encoders remain frozen, and the adaptation state is reinitialized for every input video.

\subsection{Subject-Wise Results}
\label{subsec:subject_results}

Tables~\ref{tab:biovid_subject_war}, \ref{tab:stressid_subject_war}, and \ref{tab:bah_subject_war} report the \war obtained for each of the ten target subjects. The results provide a detailed assessment of performance under subject-specific distribution shifts and reveal substantial variability across subjects and adaptation methods.

On \biovid, \ours achieves the best or tied-best performance for all ten target subjects and obtains an average \war of 81.0\%. This represents a gain of $4.9$ percentage points over T3AL, the strongest competing method with an average of 76.1\%. The improvements are particularly pronounced for Sub-4, Sub-5, Sub-6, and Sub-7, indicating that the proposed personalization strategy is effective for subjects on which the baseline methods exhibit larger performance gaps.

A similar trend is observed on \stressid, where \ours achieves the highest performance for nine of the ten subjects and improves the average \war to 81.2\%. Compared with the strongest baseline average of 75.9\%, this corresponds to an improvement of 5.3 percentage points. The gains are especially notable for Sub-1, Sub-3, Sub-5, Sub-8, and Sub-9, demonstrating improved robustness to the stronger inter-subject variability present in this dataset. Sub-4 remains the only subject for which the competing methods obtain higher performance.

On \bah, the differences between methods are smaller. Nevertheless, \ours achieves the best or tied-best result for nine subjects and obtains the highest average \war of 68.9\%, exceeding T3AL by 1.0 percentage point. The smaller overall gain reflects the greater difficulty of the dataset and the more limited separation between its expression classes. Sub-10 remains challenging, where \ours does not improve over the strongest baselines.

Overall, the subject-wise results show that the improvements of \ours are distributed across most target subjects rather than being driven by a small subset. The personalized sampled cache, together with the positive and negative target caches and the adaptive entropy and diversity gates, provides more reliable adaptation to subject-specific characteristics without updating the recognition model at inference time.

\subsection{Additional Ablation Studies}
\label{subsec:additional_ablations}

\noindent\textbf{Class prompt selection.}
The influence of class-prompt formulation is evaluated using several templates commonly adopted in CLIP-based FER and prompt-driven recognition methods~\cite{foteinopoulou2024emoclip,ni2022expanding,zhao2025enhancing}. This experiment examines whether different textual descriptions of the expression categories lead to more discriminative class embeddings and improved recognition accuracy. All prompt variants are assessed with CLIP ViT-B/32 trained on the source subjects of \biovid and tested on the corresponding target subjects. The results in Table~\ref{tab:class_prompt_ablation_full} show that performance varies noticeably across prompt formulations in terms of both \war and \fonescore. The template \emph{``a person with an expression of [CLS]''} achieves the strongest overall results and is consequently used for all prompt-based methods in the experiments.

\begin{table}[t!]
\centering
\small
\setlength{\tabcolsep}{5pt}
\renewcommand{\arraystretch}{1.1}
\begin{tabular}{lcc}
\toprule
\textbf{Prompt Template} & \textbf{\war} & \textbf{\fonescore score} \\
\midrule
a photo of a [CLS]                                      & 68.7 & 63.3 \\
a\_photo\_of\_the\_[CLS]\_face                          & 66.5 & 59.3 \\
a\_photo\_of\_one\_[CLS]\_face                          & 64.2 & 55.8 \\
a\_close-up\_photo\_of\_the\_[CLS]\_face                & 68.2 & 62.6 \\
a\_low\_resolution\_photo\_of\_a\_[CLS]\_face           & 68.7 & 64.2 \\
a\_good\_photo\_of\_a\_[CLS]\_face                      & 69.2 & 65.4 \\
a\_photo\_of\_my\_[CLS]\_face                           & 66.2 & 60.3 \\
a\_cropped\_photo\_of\_the\_[CLS]\_face                 & 64.5 & 57.5 \\
a\_photo\_of\_a\_person\_with\_[CLS]\_face              & 67.2 & 61.4 \\
\rowcolor{pinkTTA}
\textbf{a person with an expression of [CLS]}           & \textbf{69.7} & \textbf{66.6} \\
a\_portrait\_of\_a\_person\_in\_[CLS]                   & 58.2 & 46.7 \\
a\_face\_showing\_signs\_of\_[CLS]                      & 68.7 & 64.1 \\
a close up portrait of a [CLS] expression               & 69.2 & 63.0 \\
a realistic photo of a person experiencing [CLS]        & 53.5 & 40.0 \\
a cropped image of a person in [CLS]                    & 69.0 & 64.5 \\
a photo of a [CLS] person                               & 69.5 & 64.3 \\
a person with a facial expression of [CLS]              & 69.5 & 64.2 \\
a high quality photo of a [CLS] expression              & 69.5 & 64.3 \\
a photo of a face showing [CLS]                         & 69.5 & 64.3 \\
a photo of a face in [CLS]                              & 69.0 & 61.8 \\
\bottomrule
\end{tabular}
\caption{Comparison of class-prompt templates for the CLIP text encoder on \biovid with a CLIP ViT-B/32 backbone. The choice of prompt influences recognition accuracy, and \emph{``a person with an expression of [CLS]''} provides the best \war and \fonescore results.}
\label{tab:class_prompt_ablation_full}
\end{table}

\noindent\textbf{Cache construction strategies.}
Additional details are provided for the cache construction strategies compared in Table~\ref{tab:source_cache_ablation}. All variants are evaluated using the same online TTA pipeline, cache capacity, retrieval mechanism, and score-fusion rule, such that the comparison isolates the effect of cache initialization.

The \emph{classifier-based} strategy uses the $\ell_2$-normalized weights of the frozen source classifier as class representatives. Each classifier weight defines the discriminative direction associated with its corresponding class in the embedding space~\cite{yu2023source}. This approach requires no retained source samples or additional training, but provides only one fixed representative per class. The \emph{GMM-based} strategy increases the number of class representatives by treating each normalized classifier weight as the center of a Gaussian distribution and sampling multiple virtual embeddings around it~\cite{tian2021vdm}. Although this introduces diversity, the sampled features remain concentrated around a single classifier-defined class direction and may not capture the complete source feature distribution.

The \emph{generator-based} strategy trains a lightweight class-conditional feature generator using the frozen source classifier as supervision~\cite{qiu2021source}. A classification objective encourages each generated embedding to be recognized as its conditioning class, while a contrastive objective promotes intra-class compactness and inter-class separation. This produces multiple class-specific embeddings without retaining the original source samples, but requires an additional generator-training stage. The \emph{prototype-based} strategy~\cite{sharafi2026test} instead constructs class prototypes directly from embeddings of labeled source videos. These prototypes preserve information from real source subjects and therefore provide reliable class anchors, but require source-derived representations to be stored and transferred to the test stage.

\begin{table}[t!]
\centering
\resizebox{0.9\linewidth}{!}{%
\begin{tabular}{lc}
\toprule

Method & \war (\%) \\[-0.6ex]
\midrule
GMM-based                    & 74.0 \\
Classifier-based              & 77.6 \\
Generator-based                & 79.1 \\
Prototype$^{*}$~\cite{sharafi2026test}   & 81.5 \\
EEnergy-based with noise initialization & 78.2 \\ Energy-based with target initialization (ours) & 81.0 \\
\bottomrule
\end{tabular}
}

\caption{Evaluation of different personalized cache construction methods on \biovid. The prototype-based method relies on class representatives derived from labeled source data, while \ours generates class-conditioned cache entries from the current target representation through energy-based sampling. $^{*}$Uses source-derived class prototypes.}

\label{tab:source_cache_ablation}

\end{table}

\begin{table}[t!]
\centering
\small
\setlength{\tabcolsep}{6pt}
\begin{tabular}{ccc|c}
\toprule
$\lambda_s$ & $\lambda_p$ & $\lambda_n$ & \war\ (\%) \\
\midrule
0.5 & 1.0 & 1.0 & 80.1 \\
1.0 & 0.5 & 1.0 & 80.4 \\
1.0 & 1.0 & 0.5 & 80.2 \\
\rowcolor{blueTTA}
\textbf{1.0} & \textbf{1.0} & \textbf{1.0} & \textbf{81.0} \\
1.5 & 1.0 & 1.0 & 80.3 \\
1.0 & 1.5 & 1.0 & 80.5 \\
1.0 & 1.0 & 1.5 & 80.4 \\
\bottomrule
\end{tabular}

\caption{Sensitivity of \ours to the cache-fusion weights on \biovid.
Each weight is varied independently while the remaining two are fixed to
$1$. Performance remains stable across moderate changes, with
$(\lambda_s,\lambda_p,\lambda_n)=(1,1,1)$ achieving the highest \war.}
\label{tab:fusion_weight_ablation}
\end{table}

Two energy-based variants are also evaluated. In \emph{energy-based sampling with noise initialization}, each class-conditioned SGLD chain is initialized from a randomly sampled feature rather than from the current target representation. The energy defined by the frozen recognition model then guides the sample toward the corresponding class embedding. This variant requires neither stored source prototypes nor target labels, but the resulting samples are not explicitly personalized to the current video. In contrast, \emph{energy-based sampling with target initialization}, corresponding to \ours, initializes every class-conditioned chain from the current unlabeled target-video representation. Consequently, the sampled embeddings retain target-specific visual characteristics while being progressively guided toward different expression classes.

The results show that the quality and personalization of the generated embeddings substantially influence adaptation. The GMM-, classifier-, and generator-based strategies obtain 74.0\%, 77.6\%, and 79.1\% \war, respectively. Energy-based sampling initialized from noise reaches 78.2\%, indicating that class conditioning alone can generate useful cache entries but does not fully capture the current subject. The source-derived prototype strategy achieves the highest result of 81.5\%, although it requires labeled source information at test time. Without retaining source samples, labels, or prototypes, the proposed target-initialized strategy achieves a competitive 81.0\% \war. This result demonstrates that initializing the sampling process from the current target video provides effective class-specific and subject-adaptive cache representations.

\noindent\textbf{Fusion weights.}
Table~\ref{tab:fusion_weight_ablation} evaluates the sampled-, positive-,
and negative-cache weights $\lambda_s$, $\lambda_p$, and $\lambda_n$.
Each weight is varied from $0.5$ to $1.5$ while the remaining two are fixed
to $1$. Performance remains relatively stable across the evaluated settings,
indicating limited sensitivity to moderate changes in the fusion weights.
The balanced configuration
$(\lambda_s,\lambda_p,\lambda_n)=(1,1,1)$ achieves the highest \war and is
therefore used in all experiments.

\bibliography{aaai2027}
